\documentclass[conference]{IEEEtran}

\usepackage{cite}
\usepackage{amsmath,amssymb}
\usepackage{graphicx}
\usepackage{algorithm}
\usepackage{algorithmic}
\usepackage{url}
\usepackage{booktabs}
\title{Adaptive Workflow Intelligence: A Cognitive Architecture for Context-Driven Enterprise Automation}

\author{
\IEEEauthorblockN{Sreedevi Pandiyath Viswambaran}
\IEEEauthorblockA{
Georgia Institute of Technology\\
Email: sviswambaran3@gatech.edu
}
}

\begin{document}

\maketitle

\begin{abstract}
Enterprise systems increasingly rely on automated workflows, yet many AI-driven solutions remain brittle under non-stationary conditions, evolving policies, and delayed operational feedback. While reinforcement learning and large language model (LLM) agents offer partial adaptability, they do not by themselves provide persistent reflection mechanisms or straightforward integration with policy-constrained enterprise operations.

This paper introduces \emph{Adaptive Workflow Intelligence (AWI)}, a cognitive architecture for context-driven enterprise agents organized around a four-layer \emph{Perception--Cognition--Action--Reflection (PCAR)} loop. AWI treats reflection as a mechanism for continuous policy refinement and combines hybrid reasoning with reflective memory and feedback-driven adaptation to support decision making under environmental drift and operational constraints.

We evaluate AWI in a simulated enterprise decision workflow characterized by delayed outcomes and a controlled regime shift. In a drift-and-delay stress test, guardrail-constrained adaptive approaches recover more rapidly than static automation while maintaining policy compliance. Within this setting, AWI’s reflective components modestly reduce behavioral oscillation and feedback variance, illustrating the stability--agility trade-off introduced by reflective policy adaptation.
\end{abstract}


\section{Introduction}

Enterprise systems such as enterprise resource planning (ERP), procurement, and supply chain management platforms form the operational backbone of modern organizations. These systems automate large-scale workflows, enforce business policies, and ensure consistency across departments. In recent years, artificial intelligence (AI) has been increasingly integrated into such workflows to improve efficiency and decision quality. However, despite these advances, most enterprise automation remains \emph{brittle}: workflows are commonly driven by static rules or pre-trained models that struggle to adapt to evolving business contexts, regulatory changes, and delayed feedback.

A key challenge arises from the \emph{non-stationary nature of enterprise environments}. Market conditions, organizational priorities, supplier behavior, and compliance requirements evolve over time, often abruptly and under partial observability. Existing automation approaches typically require frequent manual rule updates or offline retraining to accommodate such changes, limiting scalability and responsiveness. In contrast, human decision-makers adapt naturally through experience by perceiving contextual cues, reasoning about trade-offs, acting within constraints, and reflecting on outcomes. Replicating this closed-loop, experience-driven adaptation in enterprise AI systems remains an open problem related to continual and lifelong learning in machine learning systems~\cite{thrun1998lifelong,lesort2020continual}.

Prior work has explored feedback-based control and reasoning through multiple paradigms. Autonomic computing frameworks introduced the MAPE-K (Monitor--Analyze--Plan--Execute over Knowledge) loop for self-managing systems, providing a structured feedback model for runtime adaptation~\cite{kephart2003autonomic,kephart2004mapek}. Cognitive architectures such as SOAR and ACT-R model aspects of human reasoning and memory structures in decision-making systems~\cite{laird2012soar,anderson2004actr}. While these approaches provide conceptual foundations for adaptive behavior, they do not directly address continuous learning and policy evolution in large-scale enterprise workflows.

Recent advances in large language models (LLMs) have led to agentic frameworks that interleave reasoning and action. Approaches such as ReAct combine natural language reasoning with tool use~\cite{yao2023react}, while Reflexion introduces self-evaluation loops to improve agent performance across episodes~\cite{shinn2023reflexion}. These systems demonstrate the potential of reflective reasoning but typically operate in episodic textual environments and lack persistent policy evolution grounded in structured workflow outcomes.

Reinforcement learning (RL) provides another pathway to adaptive decision-making~\cite{sutton2018rl}. Deep RL approaches such as Deep Q-Networks (DQN) and policy-gradient methods have demonstrated success in complex sequential decision problems~\cite{mnih2015dqn,schulman2017ppo}. However, standard RL formulations assume stationary reward functions and permit unrestricted exploration. These assumptions rarely hold in enterprise settings, where actions are constrained by organizational policies and errors may carry significant operational cost.

To address these limitations, we introduce \emph{Adaptive Workflow Intelligence (AWI)}, a cognitive architecture for context-driven enterprise agents that enables continuous learning within operational workflows. AWI is organized around a four-layer \emph{Perception--Cognition--Action--Reflection (PCAR)} loop that explicitly models reflection as a first-class process. Unlike reactive control loops or episodic self-correction, AWI formalizes reflection as \emph{policy evolution}, driven by discrepancies between expected and realized outcomes and supported by a persistent reflective memory. The architecture integrates hybrid reasoning---combining rules, learning-based policies, and optional LLM-based reasoning---while preserving safety, auditability, and policy compliance.

We evaluate AWI using a representative enterprise decision workflow characterized by delayed feedback, policy constraints, and a controlled regime shift. This setup allows us to examine the adaptive properties of the architecture—including recovery speed, stability, and policy compliance—without restricting the analysis to a single application domain. Experimental results show that guardrail-constrained adaptive methods recover substantially faster than static policies while maintaining full policy compliance, and that reflective memory modestly reduces oscillatory decision behavior relative to non-reflective baselines. Importantly, the goal of this work is not to introduce a new learning algorithm, but to propose an architectural framework that integrates perception, policy-constrained decision-making, workflow execution, and reflection into a unified adaptive enterprise agent loop.

\subsection{Contributions}

This paper makes the following contributions:

\begin{itemize}

\item We introduce \emph{Adaptive Workflow Intelligence (AWI)}, a cognitive architecture for enterprise automation structured around a four-layer \emph{Perception--Cognition--Action--Reflection (PCAR)} loop.

\item We propose a reflection mechanism that enables workflow policies to evolve using delayed operational feedback while maintaining policy and governance constraints typical of enterprise environments.

\item We present a controlled simulated enterprise workflow environment designed to evaluate adaptive decision-making under environmental drift and delayed rewards.

\item We empirically compare AWI against static automation and contextual bandit baselines, showing that guardrail-constrained adaptive approaches recover more rapidly from regime shifts while reflective components modestly reduce behavioral oscillation and feedback variance.

\end{itemize}

\section{Background and Related Work}

This section discusses prior work related to adaptive enterprise systems, cognitive architectures, reinforcement learning, and large language model (LLM)--based agents. We summarize their contributions and limitations and position Adaptive Workflow Intelligence (AWI) within this research landscape.

\subsection{Autonomic and Cognitive Architectures}

Early approaches to self-adaptive systems were inspired by autonomic computing principles. The MAPE-K (Monitor–Analyze–Plan–Execute over Knowledge) loop~\cite{kephart2003autonomic} introduced a feedback-driven control framework for managing complex systems through monitoring, analysis, planning, and execution cycles over shared knowledge bases~\cite{kephart2003autonomic,kephart2004mapek}. While effective for operational self-management, MAPE-K frameworks primarily rely on static knowledge representations and predefined control logic, limiting their ability to support learning-driven policy evolution in dynamic environments.

Cognitive architectures such as SOAR and ACT-R extended these ideas toward human-like reasoning by integrating declarative and procedural memory structures~\cite{laird2012soar,anderson2004actr}. These systems model perception--action cycles and symbolic reasoning processes but are primarily designed for controlled or simulated domains. As a result, they do not naturally support continuous adaptation in large-scale enterprise workflows, where decisions are constrained by policies, feedback may be delayed, and environments evolve over time.

In contrast, AWI extends feedback-based control into a learning-enabled workflow architecture that embeds perception, decision-making, execution, and reflection directly within enterprise operations. This design enables policies to evolve continuously based on observed outcomes while preserving the policy constraints and auditability required in enterprise systems.

\subsection{Reinforcement Learning in Enterprise Systems}

Reinforcement learning (RL) has been widely studied for sequential decision-making in dynamic environments~\cite{sutton2018rl}. Classical approaches such as value-based methods (e.g., Q-learning) and policy-gradient algorithms learn decision policies by optimizing cumulative reward through interaction with the environment. Deep reinforcement learning methods have demonstrated strong performance in complex sequential decision problems when sufficient exploration and relatively stable reward signals are available~\cite{mnih2015dqn,schulman2017ppo}.

However, applying RL directly in enterprise workflows presents practical challenges. Enterprise decisions are typically subject to strict policy constraints, delayed or noisy feedback, and significant operational costs when incorrect actions occur. In such settings, unrestricted exploration and slow policy convergence can lead to instability or policy violations. Furthermore, environmental conditions in enterprise systems may change over time, requiring continual adaptation rather than periodic retraining.

Contextual bandit methods provide a simplified online-learning formulation that enables faster adaptation under contextual signals while reducing exploration risk~\cite{li2010bandit}. However, standard bandit approaches typically lack mechanisms for long-term memory or reflective reasoning about historical policy shifts.

AWI differs from these approaches by combining online learning with explicit reflection and policy-constrained decision-making, enabling adaptive behavior while maintaining compliance and auditability within enterprise workflows.

\subsection{LLM-Based Agent Frameworks}

Recent advances in large language models (LLMs) have enabled agentic frameworks that interleave reasoning, acting, and reflection. Approaches such as ReAct combine natural language reasoning with tool use to enable structured interaction with external systems~\cite{yao2023react}. Similarly, Reflexion introduces a mechanism for self-evaluation and verbal feedback that allows agents to improve performance across episodes~\cite{shinn2023reflexion}. More broadly, LLM-based agents have demonstrated the potential to coordinate reasoning, planning, and tool execution in open-ended problem-solving settings.

Despite these advances, most LLM-based agents operate primarily through episodic textual interaction and do not maintain persistent, structured representations of decision policies or operational memory. Reflection is typically implemented as natural language self-critique rather than as a mechanism for quantitatively updating decision policies grounded in structured outcomes. In addition, these frameworks are generally designed for open-ended environments and do not explicitly address the policy constraints, auditability requirements, and deterministic execution models common in enterprise workflows.

AWI complements LLM-based reasoning by embedding it within a structured cognitive architecture, where perception, decision-making, execution, and reflection operate within policy-constrained workflows. In this setting, reflection influences policy evolution based on observed workflow outcomes rather than purely textual feedback.

\subsection{Summary and Research Gap}

The paradigms discussed above highlight complementary strengths but also important limitations for enterprise automation. Autonomic and cognitive architectures provide structured feedback loops and reasoning models but do not typically incorporate continuous learning mechanisms for evolving operational policies. Reinforcement learning enables optimization through interaction with dynamic environments~\cite{sutton2018rl,mnih2015dqn,schulman2017ppo}, but practical deployments must address non-stationarity, delayed feedback, and strict policy constraints. LLM-based agent frameworks demonstrate strong reasoning and reflection capabilities~\cite{yao2023react,shinn2023reflexion}, yet they are primarily designed for episodic interaction and generally lack persistent policy evolution grounded in structured enterprise outcomes.

These limitations motivate the need for an architecture that combines context-aware perception, policy-constrained decision-making, and continuous outcome-driven adaptation. Adaptive Workflow Intelligence addresses this need by integrating perception, cognition, action, and reflection into a closed-loop architecture designed to support stable, interpretable, and adaptive enterprise automation under non-stationary conditions.

\section{Adaptive Workflow Intelligence (AWI) Architecture}
\label{sec:awi_arch}

Adaptive Workflow Intelligence (AWI) is a cognitive architecture for enterprise agents designed to support continual, context-driven adaptation within operational workflows. AWI is organized around a closed-loop \emph{Perception--Cognition--Action--Reflection (PCAR)} cycle in which the system repeatedly (i) perceives enterprise context, (ii) selects a policy-constrained decision, (iii) executes workflow actions, and (iv) reflects on observed outcomes to update internal policies and memory. 

Unlike traditional workflow automation systems that rely on static rules or offline retraining, the PCAR structure embeds outcome-driven adaptation directly into the operational loop. Reflection serves as a mechanism for translating observed discrepancies between expected and realized outcomes into incremental policy updates and memory adjustments. This section describes AWI's design objectives, the PCAR loop, and the architectural mechanisms that support safe adaptation under non-stationary enterprise conditions.

\subsection{Design Objectives}
\label{subsec:design_objectives}

AWI is guided by the following design objectives:

\begin{itemize}
    \item \textbf{Context Awareness:} Represent heterogeneous enterprise signals---including structured records, workflow events, and unstructured artifacts---as a unified contextual state suitable for decision-making.

    \item \textbf{Policy-Constrained Adaptation:} Enable online learning and exploration while enforcing enterprise guardrails such as compliance rules, approval constraints, and risk thresholds.

    \item \textbf{Operational Integration and Auditability:} Connect decisions directly to executable workflow operations (e.g., APIs and task orchestration) while maintaining traceable logs of decisions, constraints, and outcomes.

    \item \textbf{Reflective Learning:} Treat reflection as a mechanism for translating outcome discrepancies into persistent updates to policies and memory, supporting continual improvement without full retraining~\cite{thrun1998lifelong,lesort2020continual}.
\end{itemize}

\subsection{PCAR Overview}
\label{subsec:pcar_overview}

\begin{figure}[t]
  \centering
  \includegraphics[width=\linewidth]{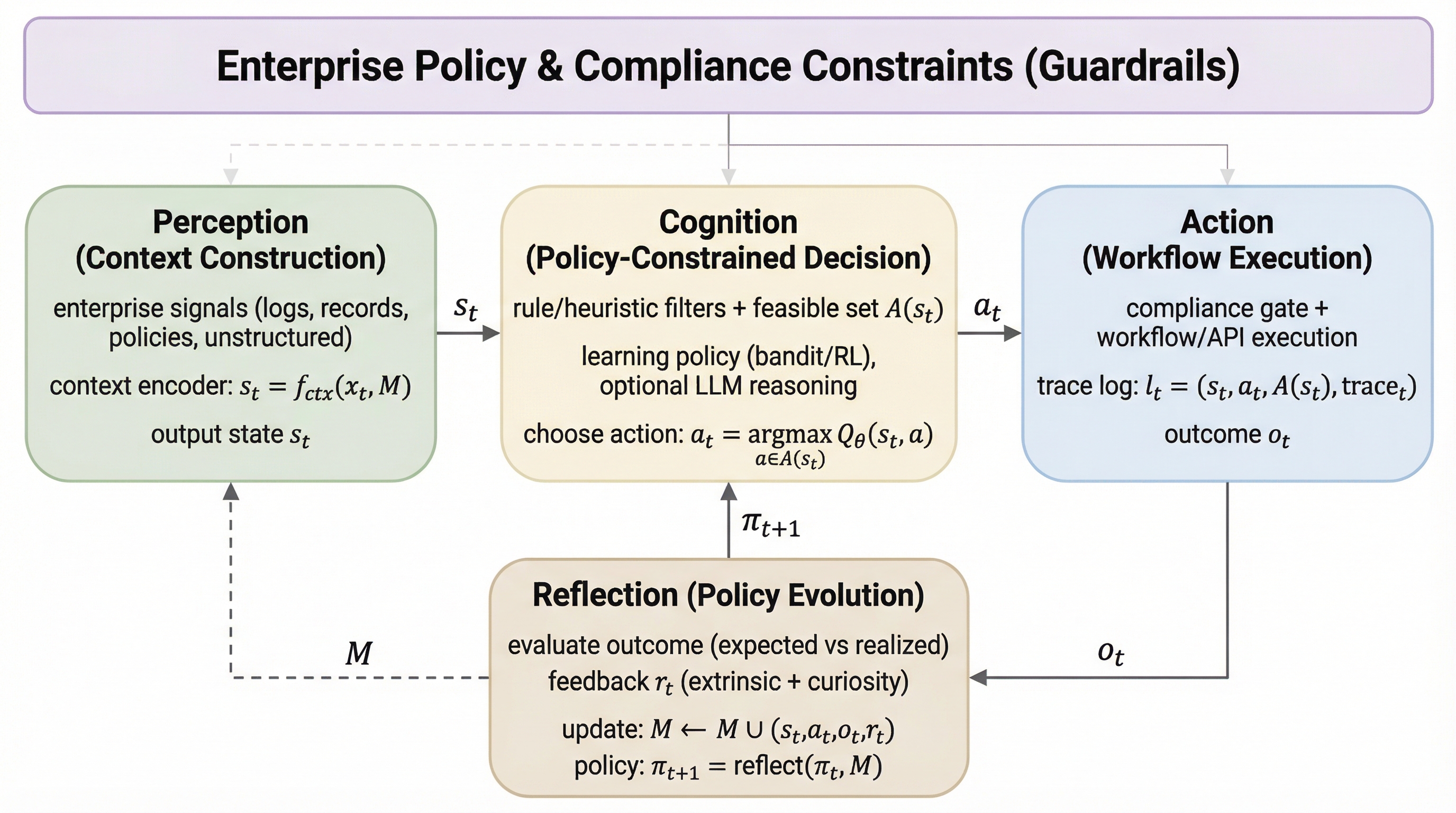}
  \caption{Adaptive Workflow Intelligence (AWI) — PCAR Loop}
  \label{fig:awi-pcar}
\end{figure}

Figure~\ref{fig:awi-pcar} conceptually illustrates the AWI PCAR loop. At each time step $t$, the Perception layer constructs the contextual state $s_t$ from enterprise observations. The Cognition layer selects an action $a_t$ subject to enterprise policy and compliance constraints. The Action layer executes the decision through workflow operations and records the resulting outcome $o_t$. 

The Reflection layer evaluates the executed action by comparing predicted and observed outcomes, producing a feedback signal $r_t$. This feedback updates reflective memory $M$ and informs a policy update that influences subsequent decisions. Through this process, reflection enables incremental policy adaptation based on observed workflow outcomes.

\begin{equation}
s_t \xrightarrow{\text{Cognition}} a_t \xrightarrow{\text{Action}} o_t \xrightarrow{\text{Reflection}} (r_t, M, \pi_{t+1}).
\end{equation}

\subsection{Perception Layer: Context Construction}
\label{subsec:perception}

The Perception layer serves as the sensing interface between enterprise environments and the agent. It ingests heterogeneous enterprise signals including workflow logs, transactional records, policy metadata, and user interactions. These inputs are normalized into a unified context state $s_t$ through a \emph{context encoder} that performs feature extraction, schema alignment, and representation construction for downstream decision modules.

Formally, let $x_t$ denote the raw enterprise observations at time $t$. The Perception layer constructs

\begin{equation}
s_t = f_{\mathrm{ctx}}(x_t),
\end{equation}

where $f_{\mathrm{ctx}}(\cdot)$ is a context encoding function that maps heterogeneous observations into a structured representation suitable for decision-making.

In practice, $s_t$ may include numerical features, categorical embeddings, policy constraint indicators, and short-term historical summaries (e.g., recent outcomes or state changes), enabling downstream components to reason over both current context and recent operational dynamics.

\subsection{Cognition Layer: Hybrid Decision-Making}
\label{subsec:cognition}

The Cognition layer maps perceived context $s_t$ to a decision $a_t$ using a hybrid reasoning stack. Enterprise workflows require both interpretability and adaptive decision-making; therefore AWI supports multiple decision modules that can be composed:

\begin{itemize}
    \item \textbf{Rule/Heuristic Module:} enforces deterministic business logic and filters actions that violate hard enterprise constraints.
    
    \item \textbf{Learning Module:} optimizes decisions under uncertainty using online learning methods such as contextual bandits~\cite{li2010bandit} or reinforcement learning techniques~\cite{sutton2018rl,mnih2015dqn,schulman2017ppo}.
    
    \item \textbf{Semantic Reasoning Module:} optionally leverages LLM-based inference to interpret unstructured context, summarize evidence, or generate candidate plans, while final decisions remain policy-gated~\cite{yao2023react,shinn2023reflexion}.
\end{itemize}

A \emph{policy manager} coordinates these modules by (i) restricting candidate actions to a feasible set $\mathcal{A}(s_t)$ defined by enterprise constraints, and (ii) ranking feasible actions using learned utility or confidence estimates:

\begin{equation}
a_t = \arg\max_{a \in \mathcal{A}(s_t)} Q_{\theta}(s_t, a),
\end{equation}

where $Q_{\theta}$ denotes a parameterized scoring function (e.g., a contextual bandit utility estimate). By separating feasibility filtering from utility maximization, the Cognition layer enables adaptive decision-making while ensuring compliance with enterprise policies.

\subsection{Action Layer: Workflow Execution and Guardrails}
\label{subsec:action}

The Action layer operationalizes selected decisions by interfacing with enterprise workflow engines, APIs, and task orchestration systems. Prior to execution, a \emph{guardrail validation step} (implemented as a compliance gate) verifies that the selected action satisfies enterprise constraints such as policy thresholds, approval requirements, and audit rules. 

Executed actions are logged together with their triggering context and constraint set to ensure traceability. Formally, the system records an execution log entry

\begin{equation}
\ell_t = (s_t, a_t, \mathcal{A}(s_t), \mathrm{trace}_t),
\end{equation}

where $\ell_t$ denotes the execution log at time $t$ and $\mathrm{trace}_t$ contains metadata such as timestamps, API responses, workflow identifiers, and validation outcomes.

After execution, the Action layer emits an outcome observation $o_t$, which may be observed immediately or after a delay depending on the workflow. Outcomes may include performance indicators, compliance results, and downstream system responses, which are subsequently processed by the Reflection layer.

\subsection{Reflection Layer: Outcome Evaluation and Policy Evolution}
\label{subsec:reflection}

The Reflection layer implements AWI's core adaptation mechanism. It evaluates executed actions by comparing predicted and observed outcomes, producing (i) a feedback signal $r_t$, (ii) updates to reflective memory $M$, and (iii) policy updates that influence subsequent decisions. Reflection therefore converts operational outcomes into incremental adjustments to the decision policy.

Policy updates can be expressed abstractly as

\begin{equation}
\pi_{t+1} = \pi_t + \eta \, \nabla_{\pi_t} J(\pi_t),
\label{eq:policy_evolution}
\end{equation}

where $\pi_t$ denotes the current decision policy, $J(\pi_t)$ represents an outcome-driven performance objective derived from observed workflow feedback, and $\eta$ is a step size controlling the magnitude of policy updates. In practice, this update may correspond to online learning rules such as contextual bandit updates or other incremental optimization procedures.

To encourage exploration in changing environments, AWI may incorporate an intrinsic curiosity signal based on prediction error~\cite{oudeyer2007curiosity,pathak2017curiosity}:

\begin{equation}
r_t = r^{\mathrm{ext}}_t + \beta 
\left\| \phi(s_{t+1}) - \hat{\phi}(s_{t+1}) \right\|_2^2,
\label{eq:curiosity}
\end{equation}

where $r^{\mathrm{ext}}_t$ denotes the external reward derived from workflow outcomes, $\phi(\cdot)$ represents the encoded next-state representation, $\hat{\phi}(\cdot)$ denotes its predicted value, and $\beta$ controls the contribution of the curiosity term. This intrinsic signal encourages exploration when prediction errors increase, enabling the system to adapt more rapidly under environmental drift.

\subsection{Reflective Memory and Knowledge Representation}
\label{subsec:memory}

AWI maintains a \emph{reflective memory} $M$ that stores experience tuples capturing context, decisions, and outcomes. Specifically, memory records

\begin{equation}
M \leftarrow M \cup \{(s_t, a_t, o_t, r_t)\}.
\end{equation}

These stored experiences support episodic recall, case-based reasoning, and trend detection (e.g., identifying emerging drift patterns). Reflective memory may also be used to condition future decision-making by augmenting the contextual state with summaries of recent outcomes or similar historical cases.

Over time, $M$ can be organized into structured representations such as enterprise knowledge graphs that link contexts, policies, and outcomes. This structure enables grounded reflection and more stable policy evolution, aligning with continual learning objectives in adaptive systems~\cite{thrun1998lifelong,lesort2020continual}.

\subsection{AWI Operational Loop}
\label{subsec:algorithm}

Algorithm~\ref{alg:awi_loop} summarizes the AWI operational cycle.

\begin{algorithm}[t]
\caption{Adaptive Workflow Intelligence (AWI) PCAR Loop}
\label{alg:awi_loop}
\begin{algorithmic}[1]
\WHILE{system active}
    \STATE Observe enterprise inputs $x_t$
    \STATE $s_t \leftarrow f_{\mathrm{ctx}}(x_t)$ \hfill (Perception)
    
    \STATE $\mathcal{A}(s_t) \leftarrow$ guardrail-constrained feasible actions
    \STATE $a_t \leftarrow \arg\max_{a \in \mathcal{A}(s_t)} Q_{\theta}(s_t, a)$ \hfill (Cognition)
    
    \STATE Execute $a_t$ through workflow APIs and log trace $\ell_t$ \hfill (Action)
    \STATE Observe outcome $o_t$ (immediate or delayed)
    
    \STATE Compute feedback $r_t$ and novelty signals \hfill (Reflection)
    \STATE Update policy parameters $\theta$
    \STATE $M \leftarrow M \cup (s_t, a_t, o_t, r_t)$ \hfill (Reflective memory)
\ENDWHILE
\end{algorithmic}
\end{algorithm}

\subsection{Summary}
\label{subsec:awi_summary}

AWI introduces a closed-loop enterprise agent architecture based on the Perception--Cognition--Action--Reflection (PCAR) cycle. By separating context construction, policy-constrained decision-making, workflow execution, and reflective policy updates, the architecture enables continual adaptation under environmental drift while maintaining enterprise compliance and auditability. The next section describes the experimental setup used to instantiate and evaluate AWI within a representative enterprise decision workflow.

\section{Methodology and Experimental Setup}
\label{sec:methodology}

This section describes how the Adaptive Workflow Intelligence (AWI) architecture is instantiated and evaluated within a representative enterprise decision workflow. The objective of the methodology is not to optimize a domain-specific application, but to examine AWI’s adaptive properties---including stability, recovery under environmental drift, policy compliance, and auditability---within a controlled yet operationally realistic setting.

\subsection{Representative Enterprise Workflow}
\label{subsec:workflow}

The evaluation is conducted in a controlled simulated enterprise workflow environment designed to capture two common operational challenges: delayed feedback and policy drift. While simplified relative to real enterprise systems, this setting enables reproducible comparison across automation strategies and isolates the effect of reflection-driven policy adaptation. The simulation models a decision workflow with delayed outcome signals and a regime shift introduced midway through execution to emulate evolving operational conditions.

We instantiate AWI using a policy-constrained enterprise decision workflow characterized by repeated evaluations over entities with evolving performance signals and delayed outcomes. While the experimental environment is implemented as a simplified simulation, the workflow structure reflects common enterprise processes involving constrained decision-making and delayed feedback.

Each decision cycle involves selecting a mitigation or control action subject to predefined organizational constraints. The experimental instantiation uses supplier risk mitigation as a concrete scenario, but the workflow abstraction is representative of a broader class of enterprise processes such as approval routing, compliance auditing, and risk governance.

Each episode corresponds to a sequence of decision steps within the workflow. At each step, the agent observes contextual signals describing the current entity state, selects an action from a constrained action set, executes the action through a simulated workflow interface, and observes an outcome after a delay. Environmental conditions are non-stationary, and the consequences of decisions may change over time, consistent with continual learning settings~\cite{thrun1998lifelong,lesort2020continual}.

\subsection{Problem Formulation and Constraints}
\label{subsec:formalism}

At each decision cycle $t$, the Perception layer constructs a contextual state $s_t$ from enterprise signals. The agent selects an action $a_t$ from a feasible action set $\mathcal{A}(s_t)$ determined by organizational policy constraints (guardrails), such that $a_t \in \mathcal{A}(s_t)$. The Action layer executes the selected decision through a workflow interface and records an auditable trace.

Outcomes are observed after a fixed delay of $k$ decision steps, producing an observation $o_{t+k}$. The Reflection layer then computes a feedback signal $r_{t+k}$ based on the discrepancy between predicted and observed outcomes. This feedback updates reflective memory $M$ and influences subsequent policy updates.

To evaluate adaptive behavior, the environment includes a drift event occurring at time $t_d$, which changes the mapping between state–action pairs and observed outcomes. This regime shift simulates evolving enterprise conditions such as changing supplier performance, updated compliance policies, or shifting operational priorities.

\subsection{Agent Configuration}
\label{subsec:agents}

The AWI system is implemented as a hierarchical multi-agent architecture consisting of a single \emph{Supervisor Agent} and multiple \emph{Worker Agents}. Each Worker Agent operates on a partition of the environment (e.g., a regional, categorical, or organizational subset) and maintains its own local policy and reflective memory.

The Supervisor Agent enforces global guardrails, coordinates policy synchronization across Workers, and monitors system-level performance signals. This hierarchical design enables scalable decision-making while preserving global policy compliance and consistent adaptation across distributed workflow contexts.

\subsection{Learning and Adaptation Mechanisms}
\label{subsec:learning}

Within each Worker Agent, the Cognition layer is implemented using an online learning policy that balances exploitation and exploration under constraints. Specifically, we employ a contextual bandit formulation~\cite{li2010bandit}, where feasible actions are ranked using context-dependent utility estimates and confidence bounds.

Action selection is restricted to the feasible action set $\mathcal{A}(s_t)$ derived from enterprise policy rules and guardrails. At each decision step, the agent selects the action that maximizes the estimated utility subject to these constraints.

Policy parameters are updated incrementally using feedback signals produced by the Reflection layer. Unlike offline retraining approaches, AWI updates policies continuously during operation, enabling rapid adaptation to environmental changes without resetting previously learned behavior.

\subsection{Reflection and Feedback Signals}
\label{subsec:reflection_impl}

Reflection evaluates the discrepancy between predicted and observed outcomes for executed actions. Feedback signals consist of two components:

\begin{itemize}
    \item \textbf{Extrinsic feedback}, derived from observed outcome changes relative to the expected workflow state (e.g., improvement or degradation in performance indicators).

    \item \textbf{Intrinsic feedback}, capturing novelty or uncertainty based on prediction error~\cite{oudeyer2007curiosity,pathak2017curiosity}.
\end{itemize}

Reflective memory $M$ stores compact summaries of recent context--action--outcome tuples and may be used to augment contextual state representations or influence decision scoring. This mechanism allows the system to incorporate recent operational experience into subsequent policy updates.

\subsection{Drift Modeling}
\label{subsec:drift}

To evaluate adaptation under non-stationary conditions, we introduce a controlled drift event during execution. At a predefined time $t_d$, the environment undergoes a regime change that alters the mapping between contextual signals and action outcomes. This shift modifies the underlying outcome distribution associated with state--action pairs, requiring the agent to adapt its policy.

Such drift events simulate disruptions commonly observed in enterprise environments, including changes in supplier performance, market conditions, or regulatory policies. Modeling controlled drift enables systematic evaluation of adaptation mechanisms commonly studied in online learning and reinforcement learning settings~\cite{sutton2018rl,mnih2015dqn,schulman2017ppo}.

\subsection{Baseline Methods and Ablations}
\label{subsec:baselines}

We compare AWI against representative automation paradigms and architectural ablations:

\begin{itemize}
    \item \textbf{Static Policy}: fixed decision rules without learning or adaptation.

    \item \textbf{Contextual Bandit (BanditOnly)}: online context-based action selection under guardrails, without reflective memory.

    \item \textbf{AWI (Full)}: contextual bandit cognition augmented with reflection-driven policy updates and reflective memory.

    \item \textbf{AWI w/o Reflection}: removes reflective memory conditioning while retaining online policy updates, isolating the effect of memory-driven reflection from standard online adaptation.

    \item \textbf{AWI Frozen Policy}: disables online updates after initialization, providing a non-adaptive control that helps distinguish conservative behavior from active adaptation.

    \item \textbf{AWI w/o Guardrails}: removes feasibility constraints, allowing the agent to select actions outside enterprise policy constraints and exposing potential compliance violations.
\end{itemize}

\subsection{Evaluation Metrics}
\label{subsec:metrics}

We report evaluation metrics aligned with enterprise operational requirements:

\begin{itemize}

\item \textbf{Policy Compliance Rate (PCR):} fraction of decision steps where the selected action satisfies enterprise guardrails,
\[
\text{PCR} = \frac{1}{T} \sum_{t=1}^{T} \mathbb{I}[a_t \in \mathcal{A}(s_t)] .
\]

\item \textbf{Auditability (Audit):} fraction of steps for which a complete execution trace is recorded, ensuring that decisions and outcomes are fully traceable.

\item \textbf{Action Switching Rate (Switch):} proportion of steps where the selected action differs from the previous step, used as a proxy for behavioral instability or oscillation.

\item \textbf{Reward Variance (VarR):} variance of the observed feedback signal $r_t$, measuring stability of delivered outcomes over time.

\item \textbf{Recovery Time (Recov):} number of steps after the drift point $t_d$ required for the rolling mean of feedback to return above a predefined fraction of pre-drift performance.
    
\end{itemize}

\subsection{Implementation Details}
\label{subsec:implementation}

The system is implemented in Python using standard numerical and machine learning libraries. The enterprise workflow environment is simulated to enable controlled experimentation while preserving key operational properties, including delayed outcomes, policy-derived feasibility constraints, and non-stationary dynamics.

All methods are evaluated under comparable computational budgets and identical environment conditions. Experiments are repeated across multiple runs with different random seeds, and reported results correspond to the mean and standard deviation of the measured metrics.

\subsection{Summary}
\label{subsec:method_summary}

This methodology instantiates AWI as a reflective, policy-constrained learning system operating within a non-stationary enterprise workflow. The experimental setup emphasizes continuous online adaptation, explicit feasibility constraints (guardrails), and outcome-driven reflection. This design provides a controlled basis for evaluating AWI’s adaptive behavior—such as stability, compliance, and recovery under drift—independently of any domain-specific optimization objective.

\section{Evaluation and Results}
\label{sec:evaluation}

This section evaluates the adaptive properties of Adaptive Workflow Intelligence (AWI) under non-stationary, policy-constrained conditions with delayed outcomes. Using the simulated enterprise workflow described in the methodology, we examine how effectively AWI adapts to environmental changes while maintaining stable, compliant, and auditable decision behavior.

\subsection{Evaluation Objectives}
\label{subsec:eval_objectives}

The evaluation is designed to answer the following questions:

\begin{itemize}
    \item Can AWI adapt to environmental drift during operation under delayed feedback?

    \item What role do feasibility constraints (guardrails) play in preserving policy compliance?

    \item How quickly do adaptive methods recover after a regime change compared with static policies?

    \item How does reflective memory influence behavioral stability and adaptation dynamics after drift?
\end{itemize}

\subsection{Evaluation Metrics}
\label{subsec:metrics_results}

The evaluation uses the metrics defined in Section~\ref{subsec:metrics}, including policy compliance rate (PCR), auditability, action switching rate (Switch), reward variance (VarR), and recovery time after drift (Recov). These metrics capture the key operational requirements of enterprise decision systems: policy compliance, auditability, behavioral stability, and adaptive recovery under non-stationary conditions.

\subsection{Experimental Setup}
\label{subsec:setup}

All methods are evaluated under identical computational budgets and environment conditions. Each experimental run consists of a horizon of 1000 decision cycles with delayed outcomes ($k=5$). A single drift event is introduced midway through execution at $t_d = 500$, altering the mapping between state–action pairs and observed outcomes. Results are averaged over five runs with different random seeds.

We compare AWI against several baseline and ablation configurations: a static rule-based policy (\textit{StaticPolicy}); an online contextual bandit baseline constrained by feasibility rules (\textit{BanditOnly})~\cite{li2010bandit}; the full AWI instantiation (\textit{AWI\_Full}); AWI without reflection-driven memory conditioning (\textit{AWI\_NoReflection}); and AWI without feasibility constraints (\textit{AWI\_NoGuardrails}).

Delayed feedback and environmental drift are known challenges for online learning and reinforcement learning systems~\cite{sutton2018rl,mnih2015dqn,schulman2017ppo}, making this setting suitable for evaluating adaptive decision architectures.

\subsection{Quantitative Results}
\label{subsec:quant_results}

Table~\ref{tab:awi_minimal_eval} summarizes performance under delayed outcomes and a controlled drift event. All guardrail-constrained methods achieve perfect policy compliance (PCR = 1.0), while removing guardrails reduces compliance and substantially increases reward variance, highlighting the importance of feasibility constraints in enterprise settings.

Compared with the static policy, adaptive methods recover substantially faster after the drift event, indicating that online learning improves post-drift recovery under delayed feedback. AWI\_Full exhibits slightly lower action-switch rates and reward variance than BanditOnly, suggesting modest gains in behavioral stability from reflective memory conditioning.

The AWI\_NoReflection configuration produces results nearly identical to BanditOnly, indicating that removing reflective memory effectively reduces the architecture to a standard contextual bandit learner operating under guardrails.

The frozen-policy ablation (AWI\_FrozenPolicy) exhibits the fastest apparent recovery and zero switching because it performs no online updates. This highlights a limitation of recovery-based metrics: conservative policies that avoid action changes may appear to recover quickly even when they do not actively adapt to environmental drift.

Overall, the results demonstrate the importance of guardrails for compliance, the benefits of online adaptation under drift, and a stability--agility trade-off between reflective memory mechanisms and purely reactive online learning.

\begin{table}[t]
\centering
\caption{Minimal evaluation under drift and delayed outcomes. Results are reported as mean $\pm$ 95\% confidence interval over 5 random seeds.}
\label{tab:awi_minimal_eval}

\resizebox{\columnwidth}{!}{
\begin{tabular}{lccccc}
\toprule
Method & PCR $\uparrow$ & Audit $\uparrow$ & Switch $\downarrow$ & VarR $\downarrow$ & Recov $\downarrow$ \\
\midrule
Static Policy       & 1.000 $\pm$ 0.000 & 0.866 $\pm$ 0.006 & 0.108 $\pm$ 0.004 & 0.043 $\pm$ 0.005 & 15.60 $\pm$ 19.42 \\
BanditOnly          & 1.000 $\pm$ 0.000 & 0.866 $\pm$ 0.006 & 0.038 $\pm$ 0.049 & 0.018 $\pm$ 0.013 & 4.20 $\pm$ 4.31 \\
AWI\_Full           & 1.000 $\pm$ 0.000 & 0.866 $\pm$ 0.006 & 0.031 $\pm$ 0.052 & 0.017 $\pm$ 0.012 & 4.40 $\pm$ 4.23 \\
AWI\_NoReflection   & 1.000 $\pm$ 0.000 & 0.866 $\pm$ 0.006 & 0.038 $\pm$ 0.049 & 0.018 $\pm$ 0.013 & 4.20 $\pm$ 4.31 \\
AWI\_NoGuardrails   & 0.972 $\pm$ 0.056 & 0.866 $\pm$ 0.006 & 0.015 $\pm$ 0.030 & 0.104 $\pm$ 0.186 & 15.40 $\pm$ 21.29 \\
AWI\_FrozenPolicy   & 1.000 $\pm$ 0.000 & 0.866 $\pm$ 0.006 & 0.000 $\pm$ 0.000 & 0.010 $\pm$ 0.003 & 1.80 $\pm$ 2.18 \\
\bottomrule
\end{tabular}
}
\vspace{2mm}

\footnotesize{
AWI\_NoReflection produces results similar to BanditOnly because removing reflective memory reduces the architecture to a standard contextual bandit learner operating under guardrails.
}
\end{table}

\subsection{Adaptation Under Drift}
\label{subsec:drift_results}

Across experimental runs, the drift event induces a transient degradation in delivered feedback, after which adaptive methods recover substantially faster than the static baseline. As shown in Table~\ref{tab:awi_minimal_eval}, the static policy exhibits a much longer recovery time, whereas adaptive approaches such as BanditOnly and AWI\_Full re-establish stable performance more quickly under delayed feedback.

Removing feasibility constraints leads to reduced policy compliance (lower PCR) and substantially higher reward variance (VarR), indicating that constraint-aware action selection is critical for safe adaptation in enterprise environments. These results highlight the importance of combining online learning with explicit enterprise guardrails when operating under non-stationary conditions.

\subsection{Ablation Analysis}
\label{subsec:ablation}

Ablation experiments isolate the contributions of key architectural components. Removing guardrails (\textit{AWI\_NoGuardrails}) reduces policy compliance (PCR) and substantially increases reward variance, highlighting the role of feasibility constraints in maintaining safe and stable operation.

Removing reflective memory (\textit{AWI\_NoReflection}) produces behavior nearly identical to the contextual bandit baseline, indicating that without reflection the architecture reduces to a standard online learning policy operating under guardrails.

The frozen-policy ablation (\textit{AWI\_FrozenPolicy}) exhibits zero action switching and the fastest apparent recovery because it performs no online updates. This result highlights a limitation of recovery-based metrics: conservative policies that avoid behavioral changes may appear to recover quickly even when they do not actively adapt to environmental drift.

Future work will expand this analysis with additional reflection and intrinsic-motivation variants~\cite{oudeyer2007curiosity,pathak2017curiosity}.

\subsection{Discussion of Results}
\label{subsec:results_discussion}

Overall, the results demonstrate three key observations. First, explicit feasibility constraints preserve policy compliance under non-stationary conditions, as evidenced by the reduced PCR and increased variance when guardrails are removed. Second, online adaptation mechanisms substantially improve post-drift recovery relative to static policies. Third, reflection-driven policy updates introduce modest improvements in behavioral stability, indicating a trade-off between stability and agility across different adaptation mechanisms. Notably, the strong recovery value reported for \textit{AWI\_FrozenPolicy} should not be interpreted as evidence of superior adaptation. Because this ablation disables online updates, it can appear stable under a recovery-based metric by avoiding behavioral changes after drift, but it does not actively adjust its policy to the new environment.

While reinforcement learning methods are a natural alternative approach for decision-making under delayed feedback and non-stationarity~\cite{sutton2018rl,mnih2015dqn,schulman2017ppo}, the present evaluation focuses on architecture-aligned baselines and ablations to isolate the effects of guardrails and reflection within AWI.

\subsection{Summary}
\label{subsec:eval_summary}

Under the drift-and-delay stress test, AWI and other adaptive methods recover substantially faster than a static policy while maintaining full policy compliance under guardrails. Removing feasibility constraints reduces compliance and increases outcome variance, underscoring the importance of constraint-aware decision-making for safe enterprise adaptation. Reflective policy updates further contribute to modest improvements in behavioral stability, highlighting a stability--agility trade-off among different adaptation mechanisms.

\section{Discussion}
\label{sec:discussion}

The evaluation highlights how AWI’s architectural separation of Perception--Cognition--Action--Reflection (PCAR), together with explicit feasibility constraints (guardrails), supports adaptive behavior in non-stationary enterprise workflows. Under the drift-and-delay stress test, adaptive approaches recover substantially faster than static rule execution while maintaining policy compliance under guardrails.

The results also emphasize the importance of constraint-aware decision making in enterprise environments. Removing feasibility constraints leads to reduced policy compliance and increased outcome variability, indicating that guardrails play a critical role in ensuring safe adaptation when automated systems interact with operational workflows.

Finally, the ablation analysis illustrates a trade-off between agility and stability among different adaptation mechanisms. Reflection-driven policy updates introduce modest improvements in behavioral stability, whereas purely reactive online learning can respond more aggressively to environmental changes. This observation suggests that architectural mechanisms such as reflective memory may serve to moderate volatility in adaptive decision systems while preserving the benefits of online learning.

\subsection{Reflection as a Mechanism for Continuous Policy Evolution}
\label{subsec:discussion_reflection}

AWI treats reflection as a first-class architectural mechanism that converts outcome discrepancies into updates to policy parameters and reflective memory. This framing differs from episodic self-correction commonly used in LLM agent frameworks~\cite{shinn2023reflexion}, as well as from purely reactive rule-based tuning mechanisms.

In AWI, reflection operates as a persistent component of the decision loop, enabling policies to evolve incrementally based on observed workflow outcomes. While the minimal evaluation primarily isolates the combined effects of online adaptation and guardrails, the architecture defines a principled location for richer reflective operators. Examples include calibrated credit assignment under delayed outcomes, memory consolidation across episodes, and controlled policy update schedules.

By separating reflection from action execution, the architecture allows these mechanisms to be incorporated without altering the operational workflow interface, supporting extensibility as enterprise decision systems evolve.

\subsection{Adaptation Under Policy Constraints}
\label{subsec:discussion_constraints}

Enterprise workflows impose hard constraints that restrict admissible actions and limit unrestricted exploration. The results show that feasibility constraints preserve policy compliance (PCR = 1.0 under guardrails), whereas removing guardrails reduces compliance and increases outcome variability. 

These findings support the architectural design choice to separate feasibility filtering (construction of the feasible action set $\mathcal{A}(s_t)$) from utility maximization in the Cognition layer. This separation enables adaptive decision-making while ensuring that selected actions remain consistent with organizational policies and audit requirements.

\subsection{Stability--Agility Trade-offs Under Delayed Feedback}
\label{subsec:discussion_tradeoffs}

Delayed outcomes introduce a stability--agility trade-off in adaptive decision systems: aggressive online updates can accelerate adjustment after environmental changes but may increase oscillatory behavior, whereas conservative updates can improve behavioral stability at the cost of slower adaptation. 

The observed differences in action-switch rates (Switch) and recovery times (Recov) across ablation configurations illustrate this trade-off in practice. In particular, reactive online learning variants respond quickly to drift, while reflection-driven mechanisms introduce modest stabilization effects by moderating update dynamics.

These observations suggest that architectural mechanisms such as reflection update cadence, confidence thresholds, and memory conditioning provide practical levers for managing stability and agility in enterprise decision systems. Future work will explore alternative reflection operators and credit-assignment strategies under longer feedback delays and multi-drift regimes.

\subsection{Multi-Agent Coordination and Scalability}
\label{subsec:discussion_multiagent}

The Supervisor/Worker hierarchy provides a deployment model in which local learning policies can specialize to environment partitions (e.g., regions or categories) while global guardrails and governance maintain organizational consistency. Such a structure aligns naturally with enterprise systems, where operational policies may vary across units but require centralized oversight and auditable change management.

Although the present evaluation focuses on a single-agent configuration to isolate architectural effects, the AWI design supports distributed deployments in which multiple Worker Agents operate over different workflow partitions under shared guardrails and supervisory coordination.

\subsection{Implications for Enterprise AI Systems}
\label{subsec:discussion_implications}

The findings suggest that effective enterprise AI systems require not only learning algorithms, but also architectural support for (i) safe operation under policy constraints, (ii) traceable and auditable execution, and (iii) continuous adaptation under non-stationary conditions and delayed feedback.

AWI addresses these requirements by embedding learning and reflection within a structured Perception--Cognition--Action--Reflection (PCAR) loop. In contrast to LLM-based reasoning frameworks that primarily operate through episodic textual interaction~\cite{yao2023react,shinn2023reflexion}, AWI grounds decisions in explicit feasibility constraints and structured workflow outcomes. This design enables policy-aware execution, auditability, and controlled adaptation in enterprise environments where unrestricted exploration or opaque reasoning may be unacceptable.

\subsection{Summary}
\label{subsec:discussion_summary}

Overall, the results emphasize that explicit guardrails are essential for maintaining policy compliance under environmental drift, and that online adaptation substantially improves post-drift recovery compared to static enterprise policies. The AWI architecture provides a structured foundation for integrating reflection-driven policy updates and memory within an auditable decision loop, supporting resilient enterprise automation in dynamic environments.

\section{Limitations and Future Work}
\label{sec:limitations}

While the current study provides evidence that AWI supports constraint-aware adaptation under drift and delayed outcomes, several limitations motivate future research.

\subsection{Simulated Environment and External Validity}
\label{subsec:lim_sim}

The evaluation is conducted in a simulated workflow environment to enable controlled drift injection, delayed feedback, and reproducible comparisons. This setting does not capture the full complexity of real enterprise operations, including heterogeneous data quality, multi-stakeholder approvals, and evolving organizational policies. Future work will validate AWI using longitudinal enterprise logs and pilot deployments, aligning evaluation with continual and lifelong learning considerations~\cite{thrun1998lifelong,lesort2020continual}.

\subsection{Feedback Design and Credit Assignment}
\label{subsec:lim_feedback}

The current prototype uses a lightweight feedback signal derived from observed outcomes under delayed effects. Real enterprise feedback may combine operational outcomes, compliance audit signals, and human feedback. Future work will study richer feedback sources, improved credit assignment under longer delays, and controlled update schedules that balance stability and agility. Intrinsic motivation mechanisms (e.g., novelty or uncertainty bonuses) are a promising extension for targeted exploration in sparse-feedback settings~\cite{oudeyer2007curiosity,pathak2017curiosity}.

\subsection{Model Expressiveness vs.\ Auditability}
\label{subsec:lim_model}

Our implementation emphasizes lightweight models to maintain interpretability and stable online updates. While suitable for a minimal evaluation, this choice may limit performance in high-dimensional enterprise contexts. Future work will evaluate more expressive representations (e.g., contextual embeddings or structured memory) while preserving auditability, constraint enforcement, and traceable policy evolution.

\subsection{Scalability and Multi-Agent Coordination}
\label{subsec:lim_scale}

Although the Supervisor/Worker structure provides a governance-aligned deployment abstraction, scaling to large numbers of agents introduces communication overhead and synchronization challenges. Future work will investigate scalable coordination strategies (e.g., periodic aggregation, asynchronous updates, and conflict resolution) under enterprise governance requirements.

\subsection{Human--AI Collaboration}
\label{subsec:lim_human}

Enterprise workflows frequently involve human approvals, overrides, and exception handling. Integrating human feedback into the Reflection layer---while maintaining traceability and avoiding destabilizing updates---is an important direction for future work. This includes incorporating structured approval outcomes, rationale capture, and human-in-the-loop safety checks.

\subsection{Summary}
\label{subsec:lim_summary}

Addressing these limitations will strengthen AWI’s applicability to real enterprise environments and improve resilient, interpretable, and policy-compliant automation under non-stationary conditions.

\section{Conclusion}
\label{sec:conclusion}

This paper introduced \emph{Adaptive Workflow Intelligence (AWI)}, a cognitive architecture for context-driven enterprise automation that supports continual adaptation under non-stationary and policy-constrained conditions. AWI is organized around a four-layer \emph{Perception--Cognition--Action--Reflection (PCAR)} loop, which treats reflection as a first-class architectural mechanism for incorporating operational outcomes into ongoing decision-making.

Using a representative enterprise workflow with delayed outcomes and a controlled drift event, we showed that online adaptive methods instantiated within the AWI framework recover substantially faster than static rule execution while maintaining policy compliance under explicit feasibility constraints. Furthermore, removing guardrails reduces compliance and increases outcome variability, underscoring the central role of constraint-aware decision making in enterprise environments.

By grounding adaptation in structured outcomes, reflective memory, and auditable execution traces, AWI provides a principled architectural foundation for building resilient, interpretable, and policy-compliant enterprise agents. While the current evaluation is conducted in a simulated workflow environment, the design targets real enterprise processes characterized by delayed feedback, evolving constraints, and limited tolerance for unsafe exploration. These properties connect naturally to broader challenges in continual and lifelong learning~\cite{thrun1998lifelong,lesort2020continual}, suggesting a path toward adaptive enterprise systems capable of learning safely from operational experience.

\bibliographystyle{IEEEtran}
\bibliography{references}

\end{document}